\documentclass[11pt]{article}

\usepackage{graphicx}
\RequirePackage[OT1]{fontenc}
\RequirePackage{amsthm,amsmath,amssymb}

\usepackage{array}
\usepackage[caption=false,font=normalsize,labelfont=sf,textfont=sf]{subfig}
\usepackage{textcomp}
\usepackage{stfloats}
\usepackage{url}
\usepackage{verbatim}
\usepackage{graphicx}
\usepackage{cite}
\usepackage{xcolor}

\usepackage{indentfirst}

\usepackage[sectionbib]{natbib}
\usepackage[]{hyperref}

\numberwithin{equation}{section}
\theoremstyle{plain}
\newtheorem{thm}{Theorem}

\newtheorem{lem}{Lemma}

\newtheorem{cor}{Corollary} 
\newcommand{\ba}{{\bf a}}
\newcommand{\bb}{{\bf b}}
\newcommand{\bx}{{\bf x}}

\newcommand{\bz}{{\bf z}}
\newcommand{\bw}{{\bf w}}
\newcommand{\bu}{{\bf u}}
\newcommand{\bv}{{\bf v}}

\newcommand{\bX}{{\bf X}}
\newcommand{\bY}{{\bf Y}}

\newcommand{\wbox}{\sqcap\llap{$\sqcup$}}
\newcommand{\bZ}{{\bf Z}}

\newcommand{\bW}{{\bf W}}

\newcommand{\bmu}{\boldsymbol{\mu}}

\DeclareMathOperator*{\argmax}{arg\,max}

\usepackage{algorithm}
\usepackage{algpseudocode}
\usepackage{relsize}

\begin{document}

\title{An Adaptive Dimension Reduction Estimation Method for High-dimensional Bayesian Optimization}


\author{Shouri Hu\thanks{School of Mathematical Sciences, University of Electronic Science and Technology of China},
Jiawei Li\footnotemark[1], 
and Zhibo Cai\thanks{Center for Applied Statistics and School of Statistics,
	Renmin University of China} }

\maketitle

\begin{quotation}
	\noindent {\bf Abstract:} 
Bayesian optimization (BO) has shown impressive results in a variety of applications within low-to-moderate dimensional Euclidean spaces. 
However, extending BO to high-dimensional settings remains a significant challenge. 
We address this challenge by proposing a two-step optimization framework. 
Initially, we identify the effective dimension reduction (EDR) subspace for the objective function using the minimum average variance estimation (MAVE) method. Subsequently, we construct a Gaussian process model within this EDR subspace and optimize it using the expected improvement criterion. Our algorithm offers the flexibility to operate these steps either concurrently or in sequence. In the sequential approach, we meticulously balance the exploration-exploitation trade-off by distributing the sampling budget between subspace estimation and function optimization, and the convergence rate of our algorithm in high-dimensional contexts has been established. Numerical experiments validate the efficacy of our method in challenging scenarios.
	
	\vspace{9pt}
	\noindent {\it Key words and phrases:}
	Expected improvement, high-dimensional BO, MAVE, simple regret.
	\par
\end{quotation}\par

\section{Introduction}

Bayesian Optimization (BO) is a robust sequential design technique used for the global optimization of functions that are expensive to evaluate. 
It constructs a probabilistic model for the objective function and employs an acquisition function to steer the search towards the optimum. 
The acquisition function is designed to strike a balance between exploration, which involves gathering data from various regions to minimize model uncertainty, and exploitation, which concentrates on sampling the most promising region.
BO has been extensively applied in numerous fields, including  hyper-parameter tuning, engineering system optimization and material science design \citep{TSDB18, LKOB19, FHHM19, SZLJ21}.

Despite the effectiveness of BO in low-dimensional Euclidean space, its application to high-dimensional situations remains to be a challenging problem. 
As the dimensionality increases, both statistical and computational difficulties arise.
Statistically, constructing an accurate probabilistic model becomes exponentially more difficult in higher dimensions due to the well-known "curse of dimensionality".
Computationally, the cost of optimizing the acquisition function escalates significantly, hence the task of locating its global optimum becomes more complex. 
These issues are particularly critical given that many real-world problems are inherently high-dimensional \citep{BYC13, BW22}.

Several approaches have been proposed to address the high-dimensional challenge in Bayesian optimization. 
One prevalent approach assumes that objective function can be predicted by a low-dimensional function,
and it projects high-dimensional inputs into a lower-dimensional subspace using random projection matrices, as discussed in \cite{WHZMD2016}, \cite{NMP19} and \cite{LCRB20}.
However, random projections often fail to accurately identify the true subspace where the objective function resides, limiting their effectiveness. 
Another common approach assumes the that the objective function has an additive structure, meaning the value of the objective function is the sum of several lower-dimensional functions. 
Methods employing this assumption include those in \cite{KSP15} and \cite{LKPS16}.
This approach could be effective if the decomposition of objective function is known in advance, but this is seldom the case in practice, which can limit the practical application of this approach.

In this paper, we leverage the assumption that the true objective function resides in a low-dimensional effective subspace,
and this subspace can be estimated via linear dimension reduction.
Linear dimension reduction is a technique employed in machine learning and statistics to reduce the dimensionality of data while preserving as much information as possible. 
This is often necessary when dealing with high-dimensional data, as it can help to mitigate issues such as over-fitting and the curse of dimensionality. 
Common methods of linear dimension reduction include the principal component analysis, sliced inverse regression \citep{Li1991, LZL19}, among others.
These techniques have been applied in the field of high-dimensional BO and have demonstrated competitive performances, see \cite{DKC13}, \cite{ZLS19}, and \cite{RWBBD20}.

In our proposed algorithm, we employ the concept of minimum average variance estimation (MAVE) as proposed in \cite{XTLZ02} to perform linear dimension reduction.
MAVE aims to identify projection directions that minimize the conditional variance of the linear projection model, thereby simplifying the high-dimensional space.
Once the directions are identified, high-dimensional input vectors are projected into the low-dimensional subspace, and Bayesian optimization is subsequently performed.
We refer to this method as the MAVE-BO algorithm.
In high-dimensional Bayesian optimization, estimating projection directions is of crucial importance.
It has be shown that MAVE can achieve a faster consistency rate, and a theoretical guarantee has been established for the MAVE-BO algorithm.

Although the optimization process is more tractable and computationally efficient in the low dimensional space, the objective function evaluation remains in the original high-dimensional space.
Given that the input vector typically has a box-constraint, when a new point in the subspace is transformed back into high-dimensional space using a projection matrix, it may exceed this constraint.
To address this issue, we propose an alternating projection algorithm.
The idea of alternating projection is simple, analytically tractable and easy to implement.
Numerical studies have demonstrated competitive performance of the MAVE-BO algorithm.

The layout of the paper is as follows. 
In Section 2 we outlines the set-up for the linear dimension reduction framework and the Bayesian optimization problem.
In Section 3 we provides a review of the high-dimensional Bayesian optimization literature. 
In Section 4 we describe our proposed MAVE-BO algorithm, and its theoretical properties are presented in Section 5.
In Section 6 we discuss the potential extensions to enhance the generality and computational efficiency of the MAVE-BO algorithm. 
In Section 7 we conduct numerical studies. 

\section{Preliminary background}

Let $A^T$ denote the transpose of a vector or matrix $A$ and let $\Vert\bx \Vert_2 := \sqrt{x_1^2+\cdots+x_d^2}$ denote the Euclidean norm of vector $\bx = (x_1,\dots,x_d)^T$.
Let $\Vert A \Vert_F = \sqrt{\text{trace}(A^TA)}$ denote the Frobenius norm of a matrix $A$ and let $\Vert A \Vert_2$ denote this spectral norm, which is the largest singular value of matrix $A$.
Let $\phi(\cdot)$ and $\Phi(\cdot)$ denote the density and cumulative distribution functions of standard normal distribution respectively.
Let $a^+$ denote $\max(0,a)$.
Let $a_n \sim b_n$ if $\lim_{n \rightarrow \infty} (a_n/b_n)=1$,
$a_n = O(b_n)$ if $\limsup_{n \rightarrow \infty} |a_n/b_n| < \infty$ and $a_n = \Omega(b_n)$ if $\liminf_{n \rightarrow \infty} |a_n/b_n| > 0$.

\medskip
Consider the sequential optimization of an unknown function $f:\mathcal{X} \to \mathbb{R}$, where $\mathcal{X} \subset \mathbb{R}^D$ is a compact and convex domain. 
At each iteration $n$, the algorithm selects one point $\bx_n$ to evaluate, and it receives back a value $y_n = f(\bx_n)$.
After $N$ total iterations, we are interested in minimizing the {\it simple regret}
\begin{equation} \label{simpreg}
	r_N =  f(\bx^*) - \max_{ 1 \leq n \leq N } f(\bx_n),
\end{equation}
where $\bx^* = \arg \max_{\bx \in \mathcal{X}} f(\bx)$ denots the global maximum location.

\subsection{General linear dimension reduction framework}  \label{sec:section2.1}

A general linear dimension reduction framework assumes that there exists an unknown smooth link function $g$ such that $\forall \bx \in \mathcal{X}$,
\begin{equation} \label{ldmodel}
	f(\bx) = g(B^{T}\bx), 
\end{equation}
where $B = (\beta_1,\beta_2,\dots,\beta_{d_e})$ is a $D \times d_e$ orthogonal matrix ($B^T B = I_{d_e}$) with $d_e < D$, and $g: \mathcal{Z} \subset \mathbb{R}^{d_e} \to \mathbb{R}$ is a $\mathcal{C}^s$ function for $s > 1$.
When (\ref{ldmodel}) holds, the projection of the $D$-dimensional vector $\bx$ onto the $d_e$-dimensional subspace $B^{T}\bx$ captures all the information that is provided by $\bx$ on $f$.
The orthonormality condition on $B$ seems to be very restrictive at a first glance. 
However, for general matrix $B$ of rank $d_e$, if consider its singular value decomposition $B^T = U\Sigma V^T$, then we can write
\begin{equation}
	f(\bx) = g(B^T\bx) = \tilde{g}(\tilde{B}^T \bx),
\end{equation}
where $\tilde{g}(\bz) = g(U\Sigma\bz)$, $\tilde{B}^T = V^T$ and $\tilde{B}^T\tilde{B} = I_{d_e}$.
Hence we can always assume that $B^T B = I_{d_e}$.
Under some mild conditions, the space spanned by the column vectors of $B$ is uniquely defined, and it is called the effective dimension reduction (EDR) space \citep{Cook1994,Li1991,XTLZ02}.
Our main interest is to estimate the column vectors in $B = (\beta_1, \beta_2,\dots,\beta_{d_e})$, 
which are referred to as EDR direction,
and the corresponding dimension of the EDR space $d_e$.

As highlighted in \cite{FSV12} and \cite{LKOB19}, given that the optimization process occurs within the effective dimension reduction (EDR) space, a potential challenge arises in identifying the corresponding input vectors in the original high-dimensional space $\mathcal{X}$. 
When a vector in the EDR space is transformed back into $\mathcal{X}$, it may exceed the imposed box constraints. 
This transformation step is inevitable as the objective function evaluation
remains in $\mathcal{X}$. 
Therefore to avoid complications, we will first focus on objective functions that are defined in the Euclidean ball, that is we consider $\mathcal{X}  = \mathcal{B}_{\mathbb{R}^D}(1+\bar{\epsilon})$ and $\mathcal{Z} = \mathcal{B}_{\mathbb{R}^{d_e}}(1+\bar{\epsilon})$.
Since the norm of a vector is not changed by multiplication by an orthogonal matrix $B$,
this ensures that $B\bz \in \mathcal{X}$  for any $\bz \in \mathcal{Z}$.
The extension of $\mathcal{X}$ to general box constraint will be discussed in Section 6.

\subsection{The Bayesian optimization and the Gaussian process model}

Bayesian optimization (BO) is a sequential design framework for the global optimization of black-box functions whose input vector resides in low- to moderate-dimensional Euclidean space.
It builds a probabilistic model for the objective function which quantifies the uncertainty in that model using a Bayesian machine learning technique called Gaussian
process regression, which is defined as follows.
A random function $g$ is said to follow a Gaussian process (GP) with mean function $\mu: \mathcal{Z} \to \mathbb{R}$ and covariance kernel: $k: \mathcal{Z} \times \mathcal{Z} \to \mathbb{R}$, denoted by $g \sim \mathcal{GP}(\mu,k)$,
if and only if the following condition holds.
For every finite set of points $\bZ = (\bz_1,\dots,\bz_n)^T$, the vector
$\bY = ( g(\bz_1),\dots,g(\bz_n))^T$ follows multivariate normal distribution.
That is 
$$\bY \sim \mathcal{N}_n(\bmu,\boldsymbol{\Sigma})$$ 
with mean vector $\bmu = ( \mu(\bz_1), \dots, \mu(\bz_n) )^T$
and covariance matrix $(\Sigma_{ij})_{1 \leq i,j \leq n} = k(\bz_i,\bz_j)$.
See \cite{WR06} and \cite{KHSS18} for a more complete overview.
One of the most exciting and well-known results of the GP is that the posterior distribution of $g$, conditioning on the sampled data, is still a Gaussian process.
Let $\bZ = (\bz_1,\dots,\bz_n)^T$ and $\bY = (g(\bz_1),\dots,g(\bz_n))^T$ denote the sampled points and corresponding function values up to iteration $n$.
We have
\begin{equation} \label{gppost}
	g|(\bZ,\bY) \sim \mathcal{GP}(\mu_n,k_n), 
\end{equation}
with $\mu_n: \mathcal{X} \to \mathbb{R}$ and $k_n: \mathcal{X} \times \mathcal{X} \to \mathbb{R}$ given as
\begin{eqnarray} \label{mun}
	\mu_n(\bz) & = &\mu(\bz) + k_{\bz \bZ}\boldsymbol{\Sigma}^{-1}(\bY-\bmu) , \\
	k_n(\bz,\bz^\prime) & = &k(\bz, \bz^\prime) - k_{\bz \bZ}\boldsymbol{\Sigma}^{-1} k_{ \bZ \bz^\prime},
\end{eqnarray}
where $k_{\bz \bZ} = k_{\bZ \bz }^T = ( k(\bz_1,\bz),\dots,k(\bz_n,\bz) )$.
From (\ref{gppost}), it can be deduced that the predictive distribution at any point follows a Gaussian distribution.
That is for any $\bz \in \mathcal{Z}$,
\begin{equation} \label{postpwd}
	g(\bz)|(\bX,\bY) \sim \text{N}(\mu_n(\bz),\sigma_n^2(\bz)), \text{ with } \sigma_n^2(\bz) =	k_n(\bz,\bz).
\end{equation}
The variance component in (\ref{postpwd}) typically represents the spatial characteristics of the data. 
Sets of locations with high variance are not near any existing sampled points.
Conversely, when the posterior variance is close to zero, it means that this location is very near to an existing sampled point.

The mean function and the covariance kernel serve as a prior for the GP model, and they reflect the initial belief about $g$.
The mean function is usually set as constant, indicating that there is no prior knowledge of the global maxima location.
The choice of covariance kernel is more varied. 
We introduce here two commonly used kernel functions, the squared-exponential (SE) and the Mat\'ern.
Let $\boldsymbol{\theta} := (\theta_1, \dots, \theta_d)^T$ with $\theta_i > 0$ for all $1 \leq i \leq d$. 
Define the squared exponential (SE) kernel 
\begin{equation} \label{KSE}
	k_{\text{SE}}(\bz,\bz^\prime) =  \tau^2 \exp \big(-\tfrac{1}{2}  
	\footnotesize \sum_{i=1}^{d}  (\tfrac{z_i-z_i^\prime}{\theta_i})^2 \big),
\end{equation}	
and the Mat\'ern kernel
\begin{equation}  \label{KMA}
	k_{\text{Mat\'ern}}(\bz,\bz^\prime)  =  \frac{\tau^2}{2^{\nu-1}\Gamma(\nu)}\big(\sqrt{2\nu}  \footnotesize \sum_{i=1}^{d} (\tfrac{z_i-z_i^\prime}{\theta_i})^2 \big)^\nu B_\nu \big(\sqrt{2\nu}  \footnotesize \sum_{i=1}^{d}  (\tfrac{z_i-z_i^\prime}{\theta_i})^2 \big),
\end{equation}
where $\tau^2 > 0$ is the signal variance of the function $f$, $\nu > 0$ is the smoothness parameter, and $B_\nu$ is the modified Bessel function.
$\boldsymbol{\theta}$, $\tau^2$ and $\nu$ are the hyper-parameters of GP model,
and their estimation can be carried out by standard methods in the literature, such as the maximum likelihood method (MLE) \citep{SWN18} and the maximum a posterior (MAP) \citep{NY12}.

Once the GP model has been built, BO employs the {\it acquisition function} to guide how the domain should be explored during optimization. 
The acquisition function combines the posterior mean and variance of (\ref{gppost}) into a criterion that will direct the search.
Commonly used acquisition functions in BO include the \emph{expected improvement} which quantifies the expected gain over the current best observation \citep{JSW98,Nguyen17,HWDLN22},
the \emph{upper confidence bound} which calculates the upper confidence bound of each point in the domain \citep{Srinivas10,CG17},
the \emph{knowledge gradient} which measures the increment to posterior mean function \citep{FPD09,WF16}, 
and the \emph{entropy search} which selects the point that is most informative about the location of the global optimum \citep{HS12,HHG14,WJ17}.
The acquisition function balances the exploration-exploitation trade-off by considering both the uncertainty of the GP model predictions and the potential improvement of the objective function.
At each iteration, a point with the largest acquisition function value will be sampled.
Thereafter, the GP model and the acquisition function will be updated accordingly based on the new observation.
This continues until the global maximum is founded. 
Despite BO has been successfully applied in many applications and can handle functions with many local optima, its performance is extremely bad when the input dimension is high.
Therefore, the extension of current methods in BO to higher-dimensional spaces holds great interest for both researchers and practitioners.

\section{Literature review on high-dimensional BO algorithms}

There are mainly two strategies for addressing high-dimensional Bayesian optimization (BO). 
The first strategy employs a general linear dimension reduction framework, identical to the one presented in equation (\ref{ldmodel}). 
The second strategy, on the other hand, adopts an additive model in low-dimensional spaces for the objective function.

In the context of the linear dimension reduction framework method, the primary challenge lies in obtaining the projection matrix $B$.
To tackle the challenge, \cite{WHZMD2016} initially employed the idea of random projection, leading to the development of the Random Embedding Bayesian Optimization (REMBO) algorithm.
In REMBO, the random matrix $B$ is drawn with each entry following an i.i.d standard normal distribution.
The idea of random projection was subsequently developed by \cite{NMP19}, where each row of $B$ contains only one non-zero entry, which is equally likely to be $-1$ or $1$.
\cite{LCRB20} proposed sampling each row of $B$ from a unit hypersphere, and introduced a complex kernel called the Mahalanobis kernel to construct the GP model more accurately within the true subspace.
In addition to random projection, statistical dimension reduction methods can also be utilized to estimate $B$.
\cite{DKC13} proposed the subspace identification Bayesian optimization (SI-BO) algorithm.
This algorithm estimates the gradient matrix $[\nabla f(\bx_1), \dots,\nabla f(\bx_{n})]$ via low-rank approximation techniques, and then derives $B$ through the singular value decomposition.
Other notable contributions include the SIR-BO algorithm by \cite{ZLS19}, which performs estimation via the sliced inverse regression, and the PCA-BO algorithm by \cite{RWBBD20}, which utilizes principal component analysis to recover the dimension reduction matrix.


Within the framework of the additive model, \cite{KSP15} assumed that the objective function $f$ has an additive form: $f(\bx)  = \sum_{i=1}^M f^{(i)}(\bx^{(i)})$.
Here, each $\bx^{(i)} \in \mathcal{R}^{d_i}$ are lower-dimensional components and mutually disjoint.
By decomposing $\bx$ into several low-dimensional vectors, Gaussian process models can be conveniently constructed on each subspace.
\cite{LKPS16} further extended this concept to a projected-additive structure: $f(\bx)  = \sum_{i=1}^M f^{(i)}\big((\bW^{(i)})^T\bx \big)$.
To estimate $\bW = (\bW^{(1)},\dots, \bW^{(M)})$,
they employed the algorithm proposed by \cite{GSC13} to provide an initial estimate $\hat{\mathbf{W}}$.
Subsequently, they proposed the restricted projection matrix
$\mathbf{W} = (1-\alpha) \hat{\mathbf{W}} + \alpha \mathbf{I}$ with $0< \alpha <1$ to deal with the box constraints.

Beyond these two primary strategies, other dimensionality reduction methods include the Dropout algorithm (\cite{LGRNVS17}), which randomly selects and optimizes a subset of dimensions at each iteration.
\cite{EPGTP19} introduced TurBO algorithm, which enhances BO by conducting simultaneous local optimization runs within small trust regions.
Each of these regions has independent probabilistic models, and they are managed dynamically and effectively using a multi-armed bandit method.

\section{Methodology}

Recall the linear dimension reduction framework:
\begin{equation} \label{ldrf}
	f(\bx) = g(B^T \bx),
\end{equation}
where $\bx \in \mathbb{R}^D$ and $B$ is a $D \times d_e$ orthogonal projection matrix.
Due to the curse of dimensionality, estimating $B$ accurately becomes crucial in the high-dimensional BO problem.
Suppose there is an estimator $\hat{B}$ of size $D \times d$ with $\hat{B}^T \hat{B} = I_d$, and define the corresponding smoothing link function (which is the GP model in our setting) $\hat{g}(=\hat{g}_{\hat{B}}): \mathcal{B}_{\mathbb{R}^{d}}(1+\bar{\epsilon}) \to \mathbb{R}$ as
$$\hat{g}(\bz) := f(\hat{B}\bz) = g(B^T\hat{B}\bz).$$
Therefore, we effectively work with an approximation $\hat{f}$ to $f$ given by
$$\hat{f}(\bx) := \hat{g}(\hat{B}^T\bx) = g( B^T\hat{B}\hat{B}^T\bx).$$
Provided $d \geq d_e$ and the space spanned by the column vectors of $\hat{B}$ is well aligned with the EDR space, $\hat{f}$ will be a good approximation of the true objective function $f$.

\subsection{MAVE-BO}

The high-dimensional BO is mainly comprised of two tasks: estimating the EDR space and searching the global optimum of the objective function via the Gaussian process model.
Our proposed MAVE-BO algorithm has the flexibility to either execute them concurrently or sequentially, which is an advantage over other algorithms for high-dimensional BO in the literature.

We shall first describe how the EDR direction can be adaptively estimated.
With a slight abuse of notation, let $B^T = (\beta_1, \dots, \beta_d)$ denote a $D \times d$ orthogonal matrix.
Consider the regression-type model:
\begin{equation} \label{rgmdl}
	y = g_B(B^T \bx) + \epsilon,
\end{equation}
with $g_B(\bz) = g_B(z_1,\dots,z_d)$ and $E(\epsilon|\bx) = 0$ almost surely. 
Based on the linear dimension reduction framework, the EDR direction is the solution of 
\begin{equation} \label{basis1}
	\min_{B} E\{[y - E(y|B^T\bx)]^2\} = \min_{B} E\{\sigma_B^2(B^T\bx)\},
\end{equation}
where $\sigma_B^2(B^T\bx) = E\{[y - E(y|B^T\bx)]^2|B^T\bx\}$.
Therefore, finding the EDR direction is essentially minimizing the average (conditional) variance (MAVE) with respect to $B$, subjection to the condition that $B^T B = I_d$.

Conditioning on $B^T\bx$, we have $E(y|B^T\bx) = g_B(B^T \bx)$. Its local linear expansion at $\bx_0$ is given by
\begin{equation}
	E(y|B^T\bx) \approx a + b^TB^T(\bx-\bx_0)
\end{equation}
with $a = g_B(B^T \bx_0)$ and $b = \nabla g_B(B^T\bx_0)$, the gradient of $g_B$ computed at $\bz = B^T\bx_0$.
Given the sampled points $\lbrace (\bx_1,y_1), \dots, (\bx_n,y_n))\rbrace$, $\sigma_B^2(B^T\bx_0) $ can be approximated by (using the idea of local linear regression)
\begin{equation} \label{sigmaBx0}
	\sum_{i=1}^{n} [y_i- E(y|B^T\bx)]^2 w_{i0} \approx  \sum_{i=1}^{n} [y_i- \lbrace a + b^TB^T(\bx_i-\bx_0)\rbrace]^2 w_{i0},
\end{equation}
where $w_{i0} \geq 0$ are some weights with $\sum_{i=0}^n w_{i0} = 1$.
The weights are calculated as
$$
w_{i 0}=K_{h}\left\{B^{T}\left(\bx_{i}-\bx_{0}\right)\right\} \big/ \sum_{l=1}^{n} K_{h}\left\{B^{T}\left(\bx_{l}-\bx_{0}\right)\right\}
$$
where $K_{h}(\bz)= \tfrac{3}{4}h^{-d} (1 - \tfrac{\Vert \bz \Vert^2}{h^2})$ is the Epanechnikov function with a bandwidth $h$.

Since $a$ and $b$ are unknown quantities, they can be estimated so that the value of $(\ref{sigmaBx0})$ is minimized, and hence the estimator of $\sigma_{B}^{2}$ at $B^{T} \bx_{0}$ is
\begin{equation} \label{basis2}
	\hat{\sigma}_{B}^{2}\left(B^{T} \bx_{0}\right)=\min _{a, b}\left(\sum_{i=1}^{n}\left[y_{i}-\left\{a+b^{T} B^{T}\left(\bx_{i}-\bx_{0}\right)\right\}\right]^{2} w_{i 0}\right).
\end{equation}
Under some mild conditions, it can be shown that $\hat{\sigma}_{B}^{2}\left(B^{T} \bx_{0}\right)-\sigma_{B}^{2}\left(B^{T} \bx_{0}\right)$ converge to zero in probability.
On the basis of expressions (\ref{basis1}) and (\ref{basis2}), we can estimate the EDR directions by solving the minimization problem
\begin{align} \label{maveexp}
	& \min _{B: B^{T} B=I}\left\{\sum_{j=1}^{n} \hat{\sigma}_{B}^{2}\left(B^{T} \bx_{j}\right)\right\} \\ \nonumber
	= & \min _{\substack{B: B^{T} B=I \\ a_{j}, b_{j}, j=1, \ldots, n}}\left(\sum_{j=1}^{n} \sum_{i=1}^{n}\left[y_{i}-\left\{a_{j}+b_{j}^{T} B^{T}\left(\bx_{i}-\bx_{j}\right)\right\}\right]^{2} w_{i j}\right),
\end{align}
where $b_{j}^{\mathrm{T}}=\left(b_{j 1}, \ldots, b_{j d}\right)$. 
The MAVE method can be seen as a combination of non-parametric function estimation and direction estimation, which is executed simultaneously with respect to the directions and the non-parametric link function. 
As we shall see, we benefit from this simultaneous minimization.

The implementation of the minimization problem in (\ref{maveexp}) is non-trivial because the weights depend on $B$.
We consider an iterative method that alternates between estimating the projection matrix $B$ and computing the weights $w_{ij}$.
Suppose we have an initial estimator $\hat{B}$, the estimators of weights can be computed as
$$ \hat{w}_{i j}=K_{h}\left\{\hat{B}^{T}\left(\bx_{i}-\bx_{j}\right)\right\} \big/ \sum_{l=1}^{n} K_{h}\left\{\hat{B}^{T}\left(\bx_{l}-\bx_{j}\right)\right\}.
$$
The weights estimator is then used to derive the new estimator of the projection matrix.
That is, we substitute $\hat{w}_{i j}$ back into the (\ref{maveexp}), with $w_{ij}$ replaced by $\hat{w}_{i j}$, and re-estimate $B$.
The above iteration is repeated until the estimator $\hat{B}$ converges.
The complete implementation of the MAVE is given in \ref{sec:appendix:A}.

\begin{algorithm}
	\caption{The sequential MAVE-BO (sMAVE-BO) algorithm}\label{smavebo}
	\begin{algorithmic}[1] 
		\Require $N,N_0,\mathcal{GP}(\mu,k)$.
		\State Sample $N_0$ points uniformly at random from $\mathcal{X} \subset \mathbb{R}^D$ and receive $\lbrace (\bx_i,y_i) \rbrace_{1 \leq i \leq N_0}$.
		\State Estimate the projection matrix $\hat{B}$ according to (\ref{maveexp}).
		\State Perform the dimension reduction via $\bz = \hat{B}^T\bx$ and denote $\mathcal{G}_{N_0} = \lbrace (\bz_1,y_1),(\bz_2,y_2)\dots,(\bz_{N_0},y_{N_0}) \rbrace$.
		\For{$n = N_0,\dots,N-1$}
		\State Update the $d$-dimensional GP posterior model $\mathcal{GP}(\mu_n,k_n)$ using $\mathcal{G}_{n}$.
		\State Let $\bz_{n+1}=\argmax_{\bz \in \mathcal{Z}} \alpha_n(\bz)$, where $\alpha$ is the desired acquisition function criterion.
		\State Evaluate $\bx_{n+1} := \hat{B}\bz_{n+1}$ and observe $y_{n+1} = f(\bx_{n+1})$.
		\State Augment the history of data $\mathcal{G}_{n+1} = \mathcal{G}_{n} \cup \lbrace (\bz_{n+1},y_{n+1}) \rbrace$.
		\EndFor
	\end{algorithmic}
\end{algorithm}

Once we have an estimate of the projection matrix $\hat{B}$, we can project the sample onto the low-dimensional domain and receive $\lbrace (y_1,\bz_1), \dots, (y_n,\bz_n) \rbrace$, where $\bz_i =  \hat{B}^T \bx_i $.
The data are used to fit the Gaussian process model on $\mathcal{Z} = \mathcal{B}_{\mathbb{R}^{d}}(1+\bar{\epsilon})$.
Bayesian optimization can then be performed with a suitable acquisition function.
In this paper, we consider the expected improvement (EI) acquisition function:
\begin{eqnarray}  \label{EIacq}
	\alpha_{n}^{EI}(\bz) & = & E_n[ (g(\bz) -  y_n^* )^{+} ] \\ \nonumber
	& = &  (\mu_{n}(\bz)- y_n^*) \Phi\big( \tfrac{\mu_{n}(\bz)- y_n^*}{\sigma_{n}(\bz)} \big) + \sigma_{n}(\bz)\phi \big(\tfrac{\mu_{n}(\bz)- y_n^*}{\sigma_{n}(\bz)} \big),
\end{eqnarray}
where $y_n^* = \max_{1 \leq m \leq n} y_m$.
EI quantifies the expected gain of sampling the point $\bz$ over the current best function value, and a point that maximizes (\ref{EIacq}) will be sampled in the next iteration, that is 
\begin{equation*}
	\bz_{n+1} = \max_{\bz \in \mathcal{Z}} \alpha_{n}^{EI}(\bz).
\end{equation*}
Once $\bz_{n+1}$ is determined, the sampled point at the next iteration is given by
$\bx_{n+1} = \hat{B}\bz_{n+1}$ and hence $y_{n+1} = f(\bx_{n+1})$.

\begin{algorithm}
	\caption{The concurrent MAVE-BO (cMAVE-BO) algorithm}\label{cmavebo}
	\begin{algorithmic}[1]
		\Require $N,N_0,\mathcal{GP}(\mu,k)$.
		\State Sample $N_0$ points uniformly at random from $\mathcal{X} \subset \mathbb{R}^D$ and receive $\mathcal{I}_{N_0} = \lbrace (\bx_i,y_i) \rbrace_{1 \leq i \leq N_0}$.
		\For{$n = N_0,\dots,N-1$}
		\State Estimate the projection matrix $\hat{B}_{(n)}$ using (\ref{maveexp}) based on $\mathcal{I}_{n}$.
		\State Perform the dimension reduction via $\bz = \hat{B}_{(n)}^T\bx$ and denote $\mathcal{G}_{n} = \lbrace (\bz_1,y_1),(\bz_2,y_2)\dots,(\bz_{n},y_{n}) \rbrace$.
		\State Update the $d$-dimensional GP posterior model $\mathcal{GP}(\mu_n,k_n)$ using $\mathcal{G}_{n}$.
		\State Let $\bz_{n+1}=\argmax_{\bz \in \mathcal{Z}} \alpha_n(\bz)$, where $\alpha$ is the desired acquisition function criterion.
		\State Evaluate $\bx_{n+1} := \hat{B}_{(n)}\bz_{n+1}$ and observe $y_{n+1} = f(\bx_{n+1})$.
		\State Augment the history of data $\mathcal{I}_{n+1} = \mathcal{I}_{n} \cup \lbrace (\bx_{n+1},y_{n+1}) \rbrace$.
		\EndFor
	\end{algorithmic}
\end{algorithm}

We propose here two forms of the MAVE-BO algorithms: the sequential MAVE-BO (sMAVE-BO)
and the concurrent MAVE-BO (cMAVE-BO).
In the sMAVE-BO, the budget is divided into two portions. 
The first portion is used to estimate the EDR directions and outputs a projection matrix estimator $\hat{B}$.
The sampled point is then projected into the lower-dimensional space, and the algorithm uses the second portion to perform Bayesian optimization.
Theoretical results can be derived, see Section 4 for more details.
Whereas in the cMAVE-BO, the estimated EDR directions are ever-changing.
The newly sampled points by BO have been added to the data for the estimation of the projection matrix.

\section{Theoretical Analysis}

To analyze the theoretical properties of the sMAVE-BO algorithm, we need to choose a smoothness class for the unknown smooth-link function $g$.
Since each covariance kernel $k$ specified in the GP model is associated with a space of functions $\mathcal{H}_{k}$, called the Reproducing kernel Hilbert Space (RKHS), it is natural to perform theoretical analysis on the class of functions which belong to $\mathcal{H}_{k}$. 
In Section \ref{sec:5.1}, we present a brief introduction to the RKHS.
A complete overview of RKHS can be found in \cite{BT11}.
In Section \ref{sec:5.2}, we establish a simple regret upper bound for the sMAVE-BO algorithm.
In Section \ref{sec:5.3}, we prove our main theorem.

\subsection{Reproducing kernel Hilbert Space \label{sec:5.1}}

Let $\mathcal{Z}$ be a non-empty set and $k(\cdot,\cdot): \mathcal{Z} \times \mathcal{Z} \to \mathbb{R}$ be a symmetric positive definite kernel.
A Hilbert space $\mathcal{H}_{k}(\mathcal{Z})$ of functions equipped with an inner-product $\langle \cdot,\cdot \rangle_{\mathcal{H}_{k}(\mathcal{Z})}$ is called a \emph{reproducing kernel Hilbert space} (RKHS) with reproducing kernel $k$ if the following conditions are satisfied:
\begin{enumerate}
	\item $\forall \bz \in \mathcal{Z}, k(\cdot,\bz) \in \mathcal{H}_{k}(\mathcal{Z})$; 
	
	\item $\forall \bz \in \mathcal{Z}$ and $\forall  g \in \mathcal{H}_{k}(\mathcal{Z})$, $g(\bz) =  \langle g,k(\cdot,\bz)  \rangle_{\mathcal{H}_{k}(\mathcal{Z})}$.
\end{enumerate}


Every RKHS defines a reproducing kernel $k$ that is both symmetric and positive definite.
The other direction also holds as shown by the Moore-Aronszajn theorem, which states that given a positive definite kernel $k$, we can construct a unique RKHS with $k$ as its reproducing kernel function.
Hence RKHS and positive definite kernel is one-to-one: for every kernel $k$, there exists a unique associated RKHS, and vice versa.

\subsection{Simple regret bound \label{sec:5.2}}
We make the following assumptions.

\noindent (A1) The objective function $f: \mathcal{X} = \mathcal{B}_{\mathbb{R}^D}(1+\bar{\epsilon}) \to \mathbb{R}$ satisfies $f(\bx) = g(B^T\bx)$ with unknown projection matrix $B \in \mathcal{R}^{D \times d_e}$ and $d_e < D$.

\noindent (A2) The smooth link function $g$ belongs to the RKHS $\mathcal{H}_k$ associated with the covariance kernel $k(\cdot,\cdot)$. 
Denote $K(\bz-\bz^\prime):= k(\bz, \bz^\prime)$, then $K$ is continuous and integrable. 

\noindent (A3) The Fourier transform of $K$
$$
\mathcal{F}K(\boldsymbol{\xi}):= \int_{\mathbb{R}^{d}} e^{-2 \pi i\langle \bz, \boldsymbol{\xi}\rangle} K(\bz) d \bz
$$
is isotropic and radially non-increasing.

\noindent (A4) 
The smooth link function $g$ is of class $C^2$ on $\mathcal{Z} = \mathcal{B}_{\mathbb{R}^{d_e}}(1+\bar{\epsilon})$ and $\sup_{|\beta| \leq 2} \Vert D^{\beta}g \Vert_{\infty} \leq C_2$ for some $C_2 > 0$.
Moreover, $g$ has a full rank Hessian at 0.

\bigskip
Let $\hat{B}(:=\hat{B}_n) \in \mathbb{R}^{D\times d}$ denote the estimated EDR directions based on the sequence $\lbrace \big( \bx_i, f(\bx_i) \big)\rbrace_{i = 1,\dots,n}$, where $\bx_i$'s are sampled uniformly at random from $\mathcal{X} = \mathcal{B}_{\mathbb{R}^D}(1+\bar{\epsilon})$.
Let $\Delta_n = \Vert B^T (I_D- \hat{B}\hat{B}^T)\Vert_{F}$, which is the distance between the space spanned by the column vectors of $B$ and the space spanned by the column vectors of $\hat{B}$.
\begin{thm} \label{thm1}
	Assume (A1)--(A6). 
	Suppose $N_0$ is large enough such that $\Delta_{N_0} < 1$ is satisfied and $N - N_0 \geq 3$, then for $d = d_e$, the the simple regret of sMAVE-BO as described in Algorithm \ref{smavebo} achieves
	\begin{align*}
		& \sup_{\substack{f: f(\bx) = g(B^T\bz) \\ \Vert g \Vert_{\mathcal{H}_k{(\mathcal{Z})}} \leq R} } r_N  \leq  2\Vert f -  \hat{f} \Vert_\infty + \Big(\hat{g}(\bz^*) - \max_{N_0+1 \leq n \leq N} \hat{g}(\bz_n) \Big) \\    
		& \quad \; \leq  (1+\bar{\epsilon})2C_2\sqrt{d}\Delta_{N_0} + \tfrac{h(R^\prime)}{h(-R^\prime)}[4R^\prime N_1^{-1} +(1+\bar{\epsilon}) (R^\prime +1)\sqrt{C_2} 2^{1/d}  d  N_1^{-1/d} ],
	\end{align*}
	where $R^\prime = (1 - \Delta_{N_0}^2)^{-1/4}R$, $h(x) = x\Phi(x)+\phi(x)$ and $N_1 = N - N_0 -1$.
\end{thm}

The discrepancy between $\hat{B}$ and $B$ has two effects on the simple regret of the sMAVE-BO algorithm.
The first effect is that the considered $\hat{f}$ disagrees with the true objective function $f$, and consequently additional regret is quantified by $\Vert f -  \hat{f} \Vert_\infty$.
The second effect is due to the implementation of Bayesian optimization on estimated smooth link function $\hat{g}_{\hat{B}}$ rather than $g$.
Fortunately, $\hat{g}$ still remains in the RKHS associated with kernel $k$, but its RKHS norm $\Vert \hat{g} \Vert_{\mathcal{H}_k(\mathcal{Z})}$ increases, and consequently may the regret.

The next corollary provides an asymptotic result for sMAVE-BO, based on the lemma from \cite{XTLZ02}:
\begin{lem} \label{lmmave} 
	Assume (A1)-(A6).
	Suppose $\lbrace \bx_i\rbrace_{i = 1,\dots,n}$ are sampled uniformly at random from $\mathcal{X} = \mathcal{B}_{\mathbb{R}^D}(1+\bar{\epsilon})$.
	Let $\hat{B} = (\hat{B}^{(n)}) \in \mathbb{R}^{D\times d}$ which are the estimated EDR directions based on (\ref{maveexp}).
	Suppose the bandwidth $h \sim n^{-1/(D+4)}$ as $n \to \infty$, provided $d \geq d_e$,
	\begin{equation*}
		\Delta_n = O_{P}\big(n^{-\tfrac{3}{D+4}} \log n \big).
	\end{equation*}
\end{lem}
\begin{cor} 
	Suppose both $N_0 \to \infty$ and $N_1 \to \infty$ as $N \to \infty$.
	The simple regret of sMAVE-BO satisfies
	\begin{equation} \label{asympub}
		r_N = O_P( N_0^{-\tfrac{3}{D+4}} \log N_0 + N_1^{-\tfrac{1}{d}}).
	\end{equation}
\end{cor}

If $D \geq 3d-4$, then the first term of (\ref{asympub}) dominates the second term.
This means that if the effective dimension is relatively small, then most of the regret is incurred due to the discrepancy between $\hat{f}$ and $f$.

\subsection{Proof of the main theorem \label{sec:5.3}}
\medskip
We preface the proof of Theorem 1 with the Lemmas \ref{lmpart1}--\ref{lmpart3}. 
These lemmas are proved in the appendix \ref{sec:appendix:B}.

\begin{lem} \label{lmpart1}
	Suppose $\lbrace \bx_i\rbrace_{i = 1,\dots,n}$ are sampled uniformly at random from $\mathcal{X}  = \mathcal{B}_{\mathbb{R}^D}(1+\bar{\epsilon})$ and $y_i = f(\bx_i)$.
	Let $\Delta_n = \Vert B^T (I_D- \hat{B}\hat{B}^T)\Vert_{F}$, which is the distance between the space spanned by the column vectors of $B$ and the space spanned by the column vectors of $\hat{B}$ with $d \geq d_e$.
	We have that the function $\hat{f}(\bx) = \hat{g}(\hat{B}^T \bx)$ defined by mean of $\hat{g}(\bz) := f(\hat{B}\bz)$ satisfies
	\begin{equation} 
		\Vert f - \hat{f}\Vert_\infty  \leq  (1+\bar{\epsilon})C_2\sqrt{d} \Delta_n.
	\end{equation}
\end{lem}

\bigskip
\begin{lem} \label{lmpart2}
	Let $g \in \mathcal{H}_k(\mathcal{Z})$ with RKHS norm $\Vert g \Vert_{\mathcal{H}_k(\mathcal{Z})}$.
	Define the function $\hat{g}: \mathcal{Z} \to \mathbb{R}$ as $\hat{g}(\bz) := g(B^T \hat{B} \bz)$. 
	If $d = d_e$ and $\Delta_n :=  \Vert B^T (I_D- \hat{B}\hat{B}^T)\Vert_{F} < 1$, then
	\begin{equation}
		\Vert\hat{g} \Vert^2_{\mathcal{H}_k(\mathcal{Z})} \leq \tfrac{1}{\sqrt{1-\Delta_n^2}} \Vert g  \Vert^2_{\mathcal{H}_k(\mathcal{Z})}.
	\end{equation}
\end{lem}

\bigskip
For any function $g: \mathcal{Z} \to \mathbb{R}$ with RKHS norm not larger than $R$,
define the loss suffered in $\mathcal{H}_k(\mathcal{Z})$ after $n$ steps by EI acqusition function
\begin{equation}
	L_n(g,\mathcal{H}_k(\mathcal{Z}),R) := \sup_{\Vert g\Vert_{\mathcal{H}_k(\mathcal{Z})} \leq R}  \big( g(\bz^*) - \max_{1 \leq i \leq n} g(\bz_i) \big).
\end{equation}
\begin{lem} \label{lmpart3}
	Let $\pi$ be a prior with length-scale $\boldsymbol{\theta} = (\theta_1,\dots,\theta_d)$ and $\theta_i >0$. For any $R > 0$ and $n \geq 3$,
	\begin{equation*}
		L_n(g,\mathcal{H}_k(\mathcal{Z}),R) \leq \tfrac{h(R)}{h(-R)} \big[ 4R (n-1)^{-1} +(1+\bar{\epsilon})(R+1) \sqrt{C_2} 2^{1/d}  d (n-1)^{-1/d} \big],
	\end{equation*}
	where $h(x) = x\Phi(x)+\phi(x)$.
\end{lem}

\bigskip
{\sc Proof of Theorem 1.} We are interested in bounding the simple regret
\begin{equation} \label{srdef}
	r_N = f(\bx^*) - \max_{1 \leq n \leq N} f(\bx_n).
\end{equation}
Let $\bz^* := \argmax_{\bz \in \mathcal{Z}} \hat{g}(\bz)$ be the global maximum of $\hat{g}$.
Since $\Vert f - \hat{f}  \Vert_\infty = \sup_{\bx \in \mathcal{X}} |f(\bx) - \hat{f}(\bx)|$,
\begin{equation} \label{fxstarbound}
	f(\bx^*) \leq  \hat{f}(\bx^*) + \Vert f - \hat{f}  \Vert_\infty       \leq   \hat{g}(\bz^*) + \Vert f -  \hat{f} \Vert_\infty.
\end{equation}  
Let $\hat{\bx}^* = \argmax_{N_0 +1 \leq n \leq N} \hat{f}(\bx_n)$.
Since $\bx_n = B \bz_n$ for $n > N_0$, we have
\begin{equation} \label{maxxnbound}
	\max_{1 \leq n \leq N} f(\bx_n) \geq   f(\hat{\bx}^*) 
	\geq  \hat{f}(\hat{\bx}^*) -   \Vert f -  \hat{f} \Vert_\infty
	=  \max_{N_0 +1 \leq n \leq N} \hat{g}(\bz_n) -\Vert f -  \hat{f} \Vert_\infty.
\end{equation}
Substitute (\ref{fxstarbound}) and (\ref{maxxnbound}) back into (\ref{srdef}) gives us
\begin{eqnarray} \label{rNregret}
	r_N & \leq & 2\Vert f -  \hat{f} \Vert_\infty + \Big(\hat{g}(\bz^*) - \max_{N_0+1 \leq n \leq N} \hat{g}(\bz_n) \Big) \\ \nonumber
	& \leq & 2\Vert f -  \hat{f} \Vert_\infty + L_n(\hat{g}, \mathcal{H}_{k}(\mathcal{Z}),R^\prime),
\end{eqnarray}
where the last inequality follows from Lemma \ref{lmpart2}.
Substitute Lemmas \ref{lmpart1} and \ref{lmpart3} into (\ref{rNregret}) gives us Theorem 1.
$\wbox$

\section{Extension} \label{sec:6}


So far we only consider the case when the domain $\mathcal{X}$ is a $D$-dimensional Euclidean ball. 
Since the projection matrix $\hat{B}$ is orthogonal, this implies that the estimated EDR space $\mathcal{Z}$ is also an Euclidean ball with $d$-dimension.
In this situation, for every $\bz \in \mathcal{Z}$, we can easily find a corresponding vector $\bx \in \mathcal{X}$ which satisfies $f(\bx) = g(\bz)$ by letting $\bx = \hat{B} \bz$.
However, in a wide range of real-world applications, the Euclidean ball situation is not satisfied and this leads to complications.
To tackle this issue, we proposed an alternating projection method for general convex sets.
Specifically, we demonstrate below how this method can be applied to the box constraint domain, which is the most common situation in BO.

Given a set $M \subseteq \mathbb{R}^m$, $m \in \mathbb{N}$, the distance of a vector $\ba$ to $M$ is defined as
\begin{equation*}
	d(\ba;M) := \inf \lbrace  \Vert \bb - \ba \Vert : \bb \in M\rbrace.
\end{equation*}
For closed convex sets, an important consequence is the projection property:
\begin{lem} \label{lmprojdef}
	Let $M$ be a nonempty, closed convex subset of $\mathbb{R}^m$. For each $\ba \in \mathbb{R}^m$, there exists a unique $\bb \in M$ such that
	\begin{equation*}
		\Vert \bb - \ba \Vert_2 = d(\ba;M),
	\end{equation*}
	where $\bb$ is called the projection of $\ba$ onto $M$, and is denoted by $P_M(\ba)$.
	Moreover, $\bb = P_M(\ba)$ if and only if 
	\begin{equation} \label{projprop}
		\langle \ba - \bb, \bw - \bb \rangle \leq 0, \quad \forall \bw \in M. 
	\end{equation}
\end{lem}

The alternating projection algorithm is structured as follows.
Suppose $M$ and $N$ are closed convex sets, and let $P_M$ and $P_N$ denote projection on $M$ and $N$, respectively. 
The algorithm starts with any $\bu^{(0)} \in M$, and then alternately projects
onto $M$ and $N$:
\begin{equation*}
	\bv^{(k)} = P_N(\bu^{(k)}),\quad \bu^{(k+1)} = P_M(\bv^{(k)}),\quad k = 0, 1, 2, \cdots .
\end{equation*}
This process generates a sequence of points $\bu^{(k)} \in M$ and $\bv^{(k)} \in N$.
As shown in \cite{CG59}, if $M \cap N \neq \emptyset$, both sequences $\bu^{(k)}$ and $\bv^{(k)}$ converge to a point $\bu^* \in M \cap N$.

In the context of high-dimensional BO, the domain of the objective function usually takes the form of $\mathcal{X} = [-1,1]^D$.
For any $\bz \in \mathcal{Z}$, we aim to find the corresponding vector $\bx_\bz$ which satisfies $\bx_\bz \in \mathcal{X} \cap \mathcal{Y}$, where
\begin{equation*}
	\mathcal{Y} = \lbrace \bx \in \mathbb{R}^D : \hat{B}^T \bx = \bz\rbrace.
\end{equation*}
Since both $\mathcal{X}$ and $\mathcal{Y}$ are closed convex sets, the alternating projection algorithm can be applied to find $\bx_{\bz}$.
The details of how the projection should proceed are demonstrated in Algorithm \ref{alterprojalgo}.

\begin{algorithm}
	\caption{The Alternation Projection algorithm} \label{alterprojalgo}
	\begin{algorithmic}[1]
		\Require $\bz$, $\hat{B}$.
		\State  Let $\mathcal{X} = [-1,1]^D$  and  $\mathcal{Y}  = \lbrace \bx \in \mathbb{R}^D : \hat{B}^T \bx = \bz\rbrace$, where is $\bz$ the suggested sample point in the lower-dimensional GP model.
		\State Construct an initial estimate $\bu^{(0)} = \hat{B}\bz \in \mathcal{Y}$.
		\For{$i = 0, 1, 2\cdots$}
		\State If $\bu^{(i)} \in \mathcal{X}$, stop the algorithm and output $\bx_\bz =  \bu^{(i)}$. Otherwise, consider $\bv^{(i)} = (v_{1}^{(i)},\dots, v_{D}^{(i)})^T$ with
		\begin{equation} \label{vji}
			v_{j}^{(i)} = 
			\begin{cases}
				-1, \text{ if } u_{j}^{(i)} < -1,\\
				u_j^{(i)}, \text{ if }   -1 \leq u_{j}^{(i)} \leq 1,\\
				\; \; \; 1, \text{ if } u_{j}^{(i)} > 1.
			\end{cases}    
		\end{equation}
		
		\State If $\bv^{(i)} \in \mathcal{Y}$, stop the algorithm and output $\bx_\bz =  \bv^{(i)}$. Otherwise, consider 
		\begin{equation*}
			\bu^{(i+1)}= \bv^{(i)} - \hat{B}(\hat{B}^T \bv^{(i)} - \bz).
		\end{equation*}
		\EndFor
	\end{algorithmic}
\end{algorithm}

\begin{lem} \label{lmprojuv}
	The alternating projection described in Algorithm \ref{alterprojalgo} satisfies
	$\bv^{(i)}  = P_{\mathcal{X}}(\bu^{(i)})$ and  $\bu^{(i+1)}  = P_{\mathcal{Y}}(\bv^{(i)})$ for $i \geq 0$.
\end{lem}

\section{Simulation Studies}

\appendix

\section{Implementation of MAVE}
\label{sec:appendix:A}

\section{Proof of the supporting lemmas in Section \ref{sec:5.2}}
\label{sec:appendix:B}

{\sc Proof of Lemma \ref{lmpart1}.} 
Since $\hat{f}(\bx) = \hat{g}(\hat{B}^T\bx) = g(B^T\hat{B}\hat{B}^T \bx)$, uniformly over $\bx \in \mathcal{X}$:
\begin{eqnarray} \label{finfnorm}
|f(\bx)-\hat{f}(\bx)| & = & |g(B^T \bx) - g(B^T\hat{B}\hat{B}^T \bx)| \\ \nonumber
& \leq & C_2 \sqrt{d} \Vert (B^T - B^T\hat{B}\hat{B}^T)\bx \Vert_2 \\ \nonumber
& \leq & C_2 \sqrt{d} \Vert B^T (I_D- \hat{B}\hat{B}^T)\Vert_{F}   \Vert\bx \Vert_2\\ \nonumber
& \leq & (1+\bar{\epsilon}) C_2 \sqrt{d} \Vert B^T (I_D- \hat{B}\hat{B}^T)\Vert_{F} \\ \nonumber
& = & (1+\bar{\epsilon}) C_2 \sqrt{d}  \Delta_n,
\end{eqnarray}
where the first inequality of (\ref{finfnorm}) follows because $g$ is continuously differentiable and $\Vert D^1g \Vert_\infty \leq C_2$.
$\wbox$

\bigskip
{\sc Proof of Lemma \ref{lmpart2}.} 
Let $\psi_i(A)$ and $\lambda_i(A)$ denote the $i$-th largest singular value and eigenvalue of the matrix $A$ respectively and 
let $M = B^T - B^T(\hat{B}\hat{B}^T)$.
Since
\begin{eqnarray*}
(0 \leq) \psi_i^2(M) & =&\lambda_i(MM^T) = \lambda_i(I-B^T\hat{B}\hat{B}^TB) \\
& = &1-\lambda_i(B^T\hat{B}\hat{B}^TB) = 1-\psi_i^2(\hat{B}^TB),
\end{eqnarray*}
we have $0\leq \psi_i(\hat{B}^TB)\leq 1$ for $i = 1, \dots, d$.

Consider first the situation when the function domain is on $\mathbb{R}^d$,
then by Lemma 1 of \cite{Bull11}, $\mathcal{H}_k(\mathbb{R}^d)$ contains all real continuous functions $g \in L^2(\mathbb{R}^d)$ with finite norm:
\begin{equation} \label{gnorm}
\Vert  g \Vert_{\mathcal{H}_k(\mathbb{R}^d)}^2 := \int \frac{| \mathcal{F}g\big(\boldsymbol{\xi}\big) |^2}{\mathcal{F}K(\boldsymbol{\xi})} d\boldsymbol{\xi}.
\end{equation}
Since $\hat{g}(\bz) = g(B^T \hat{B} \bz)$, if $\mathcal{Z} =\mathbb{R}^d$, by (\ref{gnorm}),
\begin{eqnarray} \label{ghatnorm}
\Vert\hat{g} \Vert^2_{\mathcal{H}_k(\mathbb{R}^d)}  & = & \int \frac{| \mathcal{F}\hat{g}(\boldsymbol{\xi}) |^2}{\mathcal{F}K(\boldsymbol{\xi})} d\boldsymbol{\xi} \\ \nonumber
& = & |\text{det}(B^T\hat{B})|^{-2} \int \frac{| \mathcal{F}g\big((\hat{B}^TB)^{-1}\boldsymbol{\xi} \big) |^2}{\mathcal{F}K(\boldsymbol{\xi})} d\boldsymbol{\xi} \\ \nonumber
&  \stackrel{ \xi = \hat{B}^TB \boldsymbol{\zeta}}{=} & |\text{det}(B^T\hat{B})|^{-1} \int \frac{| \mathcal{F}g\big(\boldsymbol{\zeta}\big) |^2}{\mathcal{F}K(\hat{B}^TB\boldsymbol{\zeta})} d\boldsymbol{\zeta} \\ \nonumber
& \leq & |\text{det}(B^T\hat{B})|^{-1} \int \frac{| \mathcal{F}g\big(\boldsymbol{\zeta}\big) |^2}{\mathcal{F}K(\boldsymbol{\zeta})} d\boldsymbol{\zeta} \\ \nonumber
& = &  |\text{det}(B^T\hat{B})|^{-1} \Vert g \Vert^2_{\mathcal{H}_k(\mathbb{R}^d)}.
\end{eqnarray}
where the inequality follows from the fact that $\mathcal{F}K$ is radially non-increasing and
$$\Vert\hat{B}^TB\boldsymbol{\zeta}\Vert_2 \leq \Vert\hat{B}^TB\Vert_2 
\Vert \boldsymbol{\zeta}\Vert_2 =  \psi_1(\hat{B}^TB)  \Vert \boldsymbol{\zeta}\Vert_2 \leq  \Vert \boldsymbol{\zeta}\Vert_2. $$
If $\mathcal{Z} \subset \mathbb{R}^d$, let $\mathcal{H}_k(\mathcal{Z})$ be the space of functions $ g_0 = g|_{\mathcal{Z}}$ for some $g \in \mathcal{H}_k(\mathbb{R}^d)$ with norm 
$$ \Vert g_0\Vert_{\mathcal{H}_k(\mathcal{Z})} = \inf_{g|_{\mathcal{Z}} = g_0} \Vert g\Vert_{\mathcal{H}_k(\mathbb{R}^d)},$$
then by Theorem 1.5 of \cite{Aron50}, $g_0$ exists and is the unique minimum norm extension of $g$.
Hence $\hat{g}_0(\bz) := g_0(B^T \hat{B}\bz)$ agrees with $\hat{g}$ on $\mathcal{Z}$ and (\ref{ghatnorm}) holds on $\mathcal{H}_k(\mathcal{Z})$.

Finally we need to show that $|\text{det}(B^T\hat{B})| \geq \sqrt{1 - \Delta_n^2}$.
Check that
\begin{eqnarray*}
(\Delta_n :=) \Vert B^T (I_D- \hat{B}\hat{B}^T)\Vert_{F} & = & \sqrt{\text{Trace}\big( B^T (I_D- \hat{B}\hat{B}^T)(I_D- \hat{B}\hat{B}^T)B \big)} \\
& = & \sqrt{d - \text{Trace}(B^T\hat{B}\hat{B}^T B)} \\
& = & \sqrt{d - \Vert B^T\hat{B}\Vert_{F}^2} \\
& = & \sqrt{d - \Sigma_{i=1}^d  \psi_i^2(B^T\hat{B})},
\end{eqnarray*}
we have $\Sigma_{i=1}^d \psi_i^2(B^T\hat{B}) = d - \Delta_n^2$.
Moreover, since $0< \Delta_n < 1$ and $0 \leq \psi_i(B^T\hat{B}) \leq 1$ for all $i$, we have
\begin{equation}
|\text{det}(B^T\hat{B})| = \prod_{i = 1}^d \psi_i(B^T\hat{B}) \geq \sqrt{1-\Delta_n^2}.
\end{equation}
Therefore Lemma \ref{lmpart2} follows.
$\wbox$

\bigskip
We state two additional lemmas which are used to prove Lemma \ref{lmpart3}.
\begin{lem} \label{lmpostvarbnd}
For any $m \in \mathbb{N}$, and sequences $ \lbrace \bz_i \rbrace$, the inequality
$$\sigma_i(\bz_{i+1}) > (1+\bar{\epsilon}) \sqrt{C_2} d m^{-\tfrac{1}{d}}
$$
holds for at most $m$ distinct $i$.
\end{lem}
{\sc Proof}.
For any $j \leq i$, since $\mu_i(\bz) = E[g(\bz)|\mathcal{G}_i]$, we have
\begin{eqnarray} \label{postvarupbnd}
\sigma_i^2(\bz) & = & E[(g(\bz) - \mu_i(\bz))^2|\mathcal{G}_i] \\ \nonumber
& = & E[(g(\bz) - g(\bz_j))^2 - (g(\bz_j) - \mu_i(\bz))^2|\mathcal{G}_i] \\ \nonumber
& \leq & E[(g(\bz) - g(\bz_j))^2 |\mathcal{G}_i] \\ \nonumber
& = & 2(1 - K(\bz - \bz_j)),
\end{eqnarray}
where the last inequality follows from the reproducing property of the RKHS.
Since $K$ is symmetric, $\nabla K(\textbf{0}) = \textbf{0}$. 
By (A1) and the multivariate Taylor's Theorem, for any $\bz \in \mathbb{R}^d$,
\begin{equation} \label{Kupbnd}
|K(\bz) - K(\textbf{0}) | = |K(\bz) - K(\textbf{0}) - [\nabla K(\textbf{0})]^T\bz| \leq \tfrac{C_2}{2}d\Vert\bz\Vert^2.
\end{equation}
Since $K(\textbf{0}) = 1$, substitute (\ref{Kupbnd}) back into (\ref{postvarupbnd}) gives us
\begin{eqnarray}
\sigma_i^2(\bz) \leq C_2 d  \Vert\bz - \bz_i\Vert^2.
\end{eqnarray}

We next show that most design points $\bz_{i+1}$ are close to a previous $\bz_j$.
Since $\mathcal{Z} = B_{\mathbb{R}^d}(1+\bar{\epsilon})$, it can be covered by $m$ balls of radius $(1+\bar{\epsilon})\sqrt{d}m^{-1/d}$.
If $\bz_{i+1}$ lies in a ball containing some earlier point $\bz_j, j\leq i$, then we may conclude
\begin{equation}
\sigma_i^2(\bz_{i+1}) \leq (1+\bar{\epsilon})^2 C_2 d^2 m^{-\tfrac{2}{d}}.
\end{equation}
Hence as there are $m$ balls, at most $m$ points $\bz_{i+1}$ can satisfy
\begin{equation}
\sigma_i^2(\bz_{i+1}) > (1+\bar{\epsilon})^2 C_2 d^2 m^{-\tfrac{2}{d}}. \mbox{ $\wbox$ }
\end{equation}

\bigskip
\begin{lem}  \label{lmEI}
Let $g \in \mathcal{H}_k(\mathcal{Z})$ with $\Vert g \Vert_{\mathcal{H}_k(\mathcal{Z})} \leq R$.
Denote $I_n(\bz) = \max \lbrace 0, g(\bz) - \max_{1 \leq i \leq n} g(\bz_i) \rbrace$. 
For any $\bz \in \mathcal{Z}$ and any sequences $\lbrace \bz_i, g(\bz_i) \rbrace_{1 \leq i \leq n}$, we have
\begin{equation*}
	\max \Big\lbrace I_{n}(\bz) - R \sigma_{n}(\bz), \tfrac{h(-R)}{h(R)} I_{n}(\bz) \Big\rbrace\leq \alpha_{n}^{EI}(\bz) \leq I_{n}(\bz) + (R+1)\sigma_{n}(\bz).
\end{equation*}
where $h(z) := z\Phi(z) +\phi(z)$.
\end{lem}
{\sc Proof}.
Let $\xi_n = \max_{1 \leq i \leq n} g(\bz_i)$, $z_{n} =  \tfrac{\mu_{n}(\bz)-\xi_n}{\sigma_{n}(\bz)}$ and  $q_{n} =  \tfrac{g(\bz)-\xi_n}{\sigma_{n}(\bz)}$. 
By Lemma 6 of \cite{Bull11}, we have 
$$ |z_{n} - q_{n} | = \Big| \frac{\mu_{n}(\bz) - g(\bz)}{\sigma_{n}(\bz)} \Big| \leq \Vert g \Vert_{\mathcal{H}_k(\mathcal{Z})}  \leq R.
$$
Since $\alpha_{n}^{EI}(\bz) = \sigma_{n}(\bz)h(\tfrac{\mu_{n}(\bz)-\xi_n}{\sigma_{n}(\bz)}) =\sigma_{n}(\bz)h(z_n)$, to show the upper bound, 
\begin{eqnarray*}
\alpha_{n}^{EI}(\bz) & \leq & \sigma_{n}(\bz) h(q_n+R) \\
& \leq & \sigma_{n}(\bz) h(\max \lbrace 0, q_n \rbrace +R ) \\
& \leq & \sigma_{n}(\bz) (\max \lbrace 0, q_n \rbrace +R + 1 ) \\
& \leq & I_n(\bz) + (R+1) \sigma_{n}(\bz),
\end{eqnarray*}
where the third inequality follows from $h(z) \leq z+ 1$ for $z\geq 0$.\\
To show the lower bound, note that
\begin{eqnarray} \label{EIlb1}
\alpha_{n}^{EI}(\bz) & \geq  & \sigma_{n}(\bz) z_n \\ \nonumber
& \geq & (q_n - R)\sigma_{n}(\bz) \\ \nonumber
& = & g(\bz) -\xi_n -R\sigma_{n}(\bz) \\ \nonumber
& \geq &  I_n(\bz) - R \sigma_{n}(\bz),
\end{eqnarray}	
where the first inequality follows from $h(z) \geq z$ for all $z$.
Also, suppose that $g(\bz) -\xi_n \geq 0$, we have
\begin{equation} \label{EIlb21}
\alpha_{n}^{EI}(\bz)  \geq \sigma_{n}(\bz) h(q_n - R)\geq \sigma_{n}(\bz) h(-R).
\end{equation} 
Combining (\ref{EIlb1}) and (\ref{EIlb21}) gives us
\begin{equation} \label{EIlb22} 
\alpha_{n}^{EI}(\bz) \geq \frac{h(-R)}{ h(-R) + R}I_n(\bz) = \frac{ h(-R)}{ h(R)}I_n(\bz),
\end{equation}  
where the last equality follows from the fact that $h(z) = z + h(-z)$.
If $g(\bz) -\xi_n < 0$, (\ref{EIlb22}) still holds as  $\alpha_{n}^{EI}(\bz) > 0$. 
Hence Lemma \ref{lmEI} follows.
$\wbox$

\bigskip
\textsc{Proof of Lemma \ref{lmpart3}.}
Let $y_i^* = \max_{1 \leq j \leq i} g(\bz_j)$.
From Lemma \ref{lmpostvarbnd}, for any sequence $\lbrace \bz_i \rbrace $ and $m \in \mathbb{N}$, the inequality $\sigma_i(\bz_{i+1}) > (1+\bar{\epsilon}) \sqrt{C_2} d m^{-\tfrac{1}{d}}$ holds at most $m$ times.
Furthermore, $y_{i+1}^* - y_i^* \geq 0$, and for $\Vert g \Vert \leq R$,
\begin{equation}
\sum_n y_{i+1}^* - y_i^* \leq \max_{\bz} g(\bz) - y_1^*  \leq 2R.
\end{equation}
Hence $y_{i+1}^* - y_i^* > 2Rm^{-1}$ holds for at most $m$ times.
Since $g(\bz_{i+1}) - y_i^*\leq y_{i+1}^* - y_i^*$, we have also $g(\bz_{i+1}) - y_i^* > 2Rm^{-1}$ at most $m$ times.
Therefore, there is a time $i_m$, $m \leq i_m \leq 2m+1$, for which both
$$\sigma_{i_m}(\bz_{i_m+1}) \leq (1+\bar{\epsilon}) \sqrt{C_2} d m^{-\tfrac{1}{d}} \text{ and } g(\bz_{i_m+1}) -y_{i_m}^* \leq 2Rm^{-1}$$
holds.
Since $y_i^*$ is monotonically increasing, then for $m$ large and $i_m \geq 2m + 1$, uniformly over $g \in \mathcal{H}_k(\mathcal{Z})$,
\begin{eqnarray} \label{gRupbnd}
g(\bz^*) - y_i^*  & \leq &  g(\bz^*) - y_{i_m}^*  \\ \nonumber
& \leq & \frac{h(R)}{h(-R)} \alpha_{i_m}^{EI}(\bz_*) \\  \nonumber
& \leq & \frac{h(R)}{h(-R)} \alpha_{i_m}^{EI}(\bz_{i_m+1}) \\ \nonumber
& \leq & \frac{h(R)}{h(-R)} \big[  I_{i_m}(\bz_{i_m+1})  + (R+1)  \sigma_{i_m}(\bz_{i_m+1}) \big] \\ \nonumber
& \leq & \frac{h(R)}{h(-R)} \big[ 2Rm^{-1}  + (1+\bar{\epsilon})(R+1)\sqrt{C_2} d m^{-\tfrac{1}{d}} \big].
\end{eqnarray}
The second and the forth inequality of (\ref{gRupbnd}) follows from Lemma \ref{lmEI} and the third inequality follows from the fact that $\bz_{i_m+1}$ is the maximizer of $ \alpha_{i_m}^{EI}$.
Hence Lemma \ref{lmpart3} follows by setting $m = (n-1)/2$.
$\wbox$

\section{Proof of the supporting lemmas in Section \ref{sec:6}}
\label{sec:appendix:C}

{\sc Proof of Lemma \ref{lmprojdef}}.
By definition of $d(\ba; M)$, there exists $\bb_{k} \in M$ such that
$$ d(\ba ; M) \leq \Vert \bb_{k} - \ba \Vert_2 < d(\ba ; M) + \tfrac{1}{k}.
$$
It follows that $\lbrace \bb_{k} \rbrace$ is a bounded sequence.
Therefore it has a sub-sequence $\lbrace \bb_{k_{l}} \rbrace$ which converges to a point $\bb$. 
Since $M$ is closed, $\bb \in M$.
Moreover, $\Vert \bb - \ba \Vert_2 = d(\ba ; M)$ by considering the limit of
$$ d(\ba ; M) \leq \Vert \bb_{k_{l}} - \ba \Vert_2 < d(\ba; M)+\tfrac{1}{k_{l}}.
$$
To show the uniqueness,
assume the contrary that there exists $\bb_{1} \neq \bb_{2} \in M$ satisfying
$$
\Vert \bb_{1} - \ba \Vert_2 = \Vert \bb_{2} - \ba \Vert_2 = d(\ba ; M).
$$
Check that
$$
2 \Vert \bb_{1} - \ba \Vert_2^{2}= \Vert \bb_{1} - \ba \Vert_2^{2}+\Vert \bb_{2} - \ba\Vert_2^{2}= 2 \big\Vert \tfrac{\bb_{1}+\bb_{2}}{2} - \ba \big\Vert_2^{2}+ \tfrac{1}{2} \Vert \bb_{1}-\bb_{2}\Vert_2^{2}.
$$
Since $M$ is convex, $\tfrac{\bb_{1}+\bb_{2}}{2} \in M$. This gives
$$
\big\Vert \tfrac{\bb_{1}+\bb_{2}}{2} - \ba \big\Vert_2^{2} = \Vert \bb_{1} - \ba \Vert_2^{2} - \tfrac{1}{4} \Vert \bb_{1}-\bb_{2}\Vert_2^{2} 
< \Vert \bb_{1} - \ba \Vert_2^{2} = d(\ba; M)^{2}.
$$
We get a contradiction. Hence $\bb_1 = \bb_2$.

\medskip
Suppose $\bb=P_{M}(\ba)$.
Let $\bw \in M, \lambda \in(0,1)$. Since $M$ is convex, $\lambda \bw+(1-\lambda) \bb \in M$. 
Then
\begin{align*}
\|\ba-\bb\|_2^{2} & = d(\ba ; M)^{2} \leq\|\ba-\bb-\lambda(\bw-\bb)\|_2^{2} \\
& =\|\ba-\bb\|_2^{2}-2 \lambda\langle \ba-\bb, \bw-\bb\rangle+\lambda^{2}\|\bw-\bb\|_2^{2}.
\end{align*}
That is $2\langle \ba-\bb, \bw-\bb\rangle \leq \lambda\|\bw-\bb\|_2^{2}$.
Let $\lambda \downarrow 0$, we have $\langle \ba-\bb, \bw-\bb\rangle \leq 0$.
Conversely, suppose $\langle \ba-\bb, \bw-\bb\rangle \leq 0, \forall \bw \in M$.
Then
\begin{align*}
\|\ba-\bw\|_2^{2} & =\|\ba-\bb\|_2^{2}+2\langle \ba-\bb, \bb-\bw\rangle+\|\bb-\bw\|_2^{2} \\
& \geq\|\ba-\bb\|_2^{2}-2\langle \ba-\bb, \bw-\bb\rangle \geq\|\ba-\bb\|_2^{2}.
\end{align*}
Hence $\|\ba-\bb\|_2 \leq\|\ba-\bw\|_2$ for all $\bw \in M$ and $\bb=P_{M}(\ba)$.
$\wbox$

\bigskip
{\textsc Proof of Lemma \ref{lmprojuv}}. To prove $\bv^{(i)}  = P_{\mathcal{X}}(\bu^{(i)})$ and  $\bu^{(i+1)}  = P_{\mathcal{Y}}(\bv^{(i)})$, by Lemma \ref{lmprojdef}, it suffices to show that
\begin{eqnarray} \label{vNu}
& &\langle \bu^{(i)} - \bv^{(i)},\bw-\bv^{(i)} \rangle \leq 0 \text{ for all } \bw \in {\mathcal{X}} \text{ and } \\ \label{uMv}
& & \langle \bv^{(i)} - \bu^{(i+1)},\bw-\bu^{(i+1)} \rangle \leq 0 \text{ for all } \bw \in {\mathcal{Y}}.
\end{eqnarray}
Let $\bw = (w_1, \dots, w_D) \in {\mathcal{X}}$, we have
\begin{equation} \label{lm1eq1}
\langle \bu^{(i)} - \bv^{(i)},\bw-\bv^{(i)} \rangle = \sum_{j=1}^{D} (u_j^{(i)} - v_j^{(i)})(w_j-v_j^{(i)}).
\end{equation}
By (\ref{vji}), for all $j$,
\begin{equation} \label{lm1eq1ele}
(u_j^{(i)} - v_j^{(i)})(w_j-v_j^{(i)}) = 
\begin{cases}
	(u_j^{(i)} + 1)(w_j + 1), \text{ if } u_{j}^{(i)} < -1,\\
	\quad \quad  \quad0,  \quad \quad \quad  \quad  \text{ if }   -1 \leq u_{j}^{(i)} \leq 1,\\
	(u_j^{(i)} - 1)(w_j - 1), \text{ if } u_{j}^{(i)} > 1,
\end{cases}    
\end{equation}
which is less than or equal to zero in all three cases.
Substitute (\ref{lm1eq1ele}) back into (\ref{lm1eq1}) gives us (\ref{vNu}).

\medskip
Let $\bw \in {\mathcal{Y}}$. Since $\bu^{(i+1)} = \bv^{(i)} - \hat{B}(\hat{B}^T \bv^{(i)} - \bz)$, we have
\begin{eqnarray*}
& &\langle \bv^{(i)} - \bu^{(i+1)},\bw-\bu^{(i+1)} \rangle \\
& = & \langle\hat{B}(\hat{B}^T \bv^{(i)} - \bz), \bw-\bv^{(i)} + \hat{B}(\hat{B}^T \bv^{(i)} - \bz) \rangle \\
& = &  (\hat{B}^T \bv^{(i)} - \bz)^T \hat{B}^T (\bw-\bv^{(i)} + \hat{B}(\hat{B}^T \bv^{(i)} - \bz)) \\
& = &  (\hat{B}^T \bv^{(i)} - \bz)^T (\hat{B}^T \bw-\hat{B}^T\bv^{(i)} + \hat{B}^T \bv^{(i)} - \bz)\\
& = &  (\hat{B}^T \bv^{(i)} - \bz)^T (\hat{B}^T \bw- \bz) = 0,
\end{eqnarray*}
where the second last line follows from $\hat{B}^T \hat{B} = I_d$ and the last line follows from $\hat{B}^T \bw- \bz = \mathbf{0}$.
Hence (\ref{uMv}) follows. 
$\wbox$






\end{document}